\documentclass[letterpaper, 10 pt, conference]{ieeeconf}  

\IEEEoverridecommandlockouts                              

\usepackage{multicol}
\usepackage{graphicx}
\usepackage{amsmath}
\usepackage{bbm}
\usepackage{dblfloatfix}
\usepackage[ruled,linesnumbered,noend]{algorithm2e}

\let\labelindent\relax
\usepackage{enumitem}

\usepackage[zerostyle=c,scaled=.94]{newtxtt}
\SetCommentSty{mycommfont}
\SetFuncSty{myprocname}

\usepackage{mathrsfs}
\usepackage{bm}
\usepackage[makeroom]{cancel}
\usepackage{amssymb}
\usepackage{upgreek}
\usepackage{multirow}
\usepackage{amsfonts}
\usepackage{here}
\usepackage[normalem]{ulem}
\usepackage{mdframed}
\usepackage{flushend}

\usepackage{booktabs}
\usepackage{color}
\usepackage{subcaption}
\usepackage[font=small, skip=5pt]{caption}
\usepackage{listings}
\usepackage{hyperref}
\hypersetup{
    colorlinks,
    linkcolor={red!50!black},
    citecolor={blue!80!black},
    urlcolor={blue!80!black}
}
\usepackage{wrapfig}
\usepackage{pdflscape}
\usepackage{setspace}
\makeatletter
\let\NAT@parse\undefined
\makeatother
\usepackage[numbers]{natbib}

\makeatletter
\patchcmd{\@algocf@start}
  {-1.5em}
  {0pt}
  {}{}
\makeatother

\newcounter{mylabelcounter}
\makeatletter
\newcommand{\labelText}[2]{%
#1\refstepcounter{mylabelcounter}%
\immediate\write\@auxout{%
  \string\newlabel{#2}{{1}{\thepage}{{\unexpanded{#1}}}{mylabelcounter.\number\value{mylabelcounter}}{}}%
}%
}
\makeatother

\usepackage{cleveref}
\usepackage{tikz}
\DeclareMathAlphabet{\mathcal}{OMS}{cmsy}{m}{n}
\DeclareMathAlphabet\mathbfcal{OMS}{cmsy}{b}{n}

\newcommand{\E}{\mathbb{E}}

\graphicspath{{figs/}}

\newcommand{\eg}{\textit{e.g.}}
\newcommand{\ie}{\textit{i.e.}}

\newcommand{\Sspace}{\mathcal{S}}
\newcommand{\Aspace}{\mathcal{A}}
\newcommand{\Ospace}{\mathcal{O}}

\newcommand{\Val}{\textsc{Val}}
\newcommand{\PriorVal}{\textsc{PriorVal}}
\newcommand{\Norm}{\textsc{Norm}}

\newcommand{\rmax}{R_{\text{max}}}
\newcommand{\rmin}{R_{\text{min}}}
\newcommand{\SO}{\text{SO}}

\usepackage{xcolor,colortbl}
\definecolor{Gray}{gray}{0.85}
\definecolor{LightCyan}{rgb}{0.88,1,1}
\newcolumntype{a}{>{\columncolor{Gray}}c}
\newcolumntype{b}{>{\columncolor{white}}c}

\definecolor{purple}{rgb}{0.37, 0.18, 0.67}
\definecolor{darkgreen}{rgb}{0.08, 0.55, 0.08}
\definecolor{blue1}{rgb}{0.2, 0.2, 0.6}
\definecolor{ltgreen}{rgb}{0.5, 0.8, 0.5}
\definecolor{green1}{rgb}{0.2, 0.6, 0.2}
\definecolor{green2}{rgb}{0.1, 0.4, 0.1}
\definecolor{dkgreen}{rgb}{0,0.7,0}
\definecolor{dkgreen2}{rgb}{0,0.55,0}
\definecolor{dkgreen3}{rgb}{0,0.4,0}
\definecolor{gray}{rgb}{0.5,0.5,0.5}
\definecolor{mauve}{rgb}{0.58,0,0.82}
\definecolor{magenta}{rgb}{0.82,0,0.82}

\newcommand{\omark}[2][]{{\color{dkgreen3}
    \ifthenelse{\equal{#1}{}}{}{\textsf{[#1]}:}#2}}

\definecolor{purple}{rgb}{0.858, 0.08, 0.85}
\definecolor{darkgreen}{rgb}{0.08, 0.55, 0.08}
\definecolor{blue1}{rgb}{0.2, 0.2, 0.6}
\definecolor{green1}{rgb}{0.2, 0.6, 0.2}
\definecolor{green2}{rgb}{0.1, 0.4, 0.1}

\title{\LARGE \bf
  A System for Generalized 3D Multi-Object Search
}

\author{Kaiyu Zheng, Anirudha Paul, Stefanie Tellex
  \thanks{Department of Computer Science, Brown University, Providence, RI, USA.
    This work is supported by the US Army under grant number W911NF2120296, and AFOSR under grant number FA9550-21-1-0214.
  Corresponding author: Kaiyu Zheng (\{\texttt{kzheng10@cs.brown.edu}\}).}}%

\begin{document}

\maketitle
\thispagestyle{empty}
\pagestyle{empty}

\begin{abstract}
  Searching for objects is a fundamental skill for robots. As such, we expect object search to eventually become an off-the-shelf capability for robots, similar to \eg, object detection and SLAM. In contrast, however, no system for 3D object search exists that generalizes across real robots and environments. In this paper, building upon a recent theoretical framework that exploited the octree structure for representing belief in 3D, we present GenMOS (Generalized Multi-Object Search), the first general-purpose system for multi-object search (MOS) in a 3D region that is robot-independent and environment-agnostic. GenMOS takes as input point cloud observations of the local region, object detection results, and localization of the robot's view pose, and outputs a 6D viewpoint to move to through online planning.
  In particular, GenMOS uses point cloud observations in three ways: (1) to simulate occlusion; (2) to inform occupancy and initialize octree belief; and (3) to sample a belief-dependent graph of view positions that avoid obstacles.
  We evaluate our system both in simulation and on two real robot platforms. Our system enables, for example, a Boston Dynamics Spot robot to find a toy cat hidden underneath a couch in under one minute. We further integrate 3D local search with 2D global search to handle larger areas, demonstrating the resulting system in a 25m$^2$ lobby area.

\end{abstract}

\section{Introduction}
The ability to search is a valuable skill for robots both by itself (\eg, for search and rescue~\cite{nourbakhsh2005human,kamegawa2020development}) and as a prerequisite for downstream tasks involving the target objects \cite{sharma2021skill, saycan2022arxiv}.
However, in contrast to other basic capabilities such as object detection, SLAM, and motion planning, to date, there has been no general-purpose object search system or package available for the robotics community, and robot platforms are yet to come equipped with the ability to search.
This paper aims to enable most robots today with a movable viewport to perform object search in an off-the-shelf manner; we call this problem \emph{generalized object search}.

Developing such a system is non-trivial, as object search in the real world requires reasoning under partial observability at 3D scale, subject to limited field-of-view (FOV), occlusion, and unreliable object detectors. Furthermore, such a system should allow interpretation of the robot's state of uncertainty and search behavior. Previous work on object search modeled the problem as a Partially Observable Markov Decision Process (POMDP)~\cite{kaelbling1998planning}, which captures key aspects of uncertainty in object search, yet constrained the problem in 2D for computational feasibility \cite{wandzel2019multi, zeng2020semantic}. Other work attempted to learn end-to-end search policies given visual input \cite{zhu2017target,batra2020objectnav}; nevertheless, those methods were primarily evaluated in simulation and generalization across different real robots and environments remains extremely difficult.

\begin{figure}[t]
  \centering
  \includegraphics[width=\linewidth]{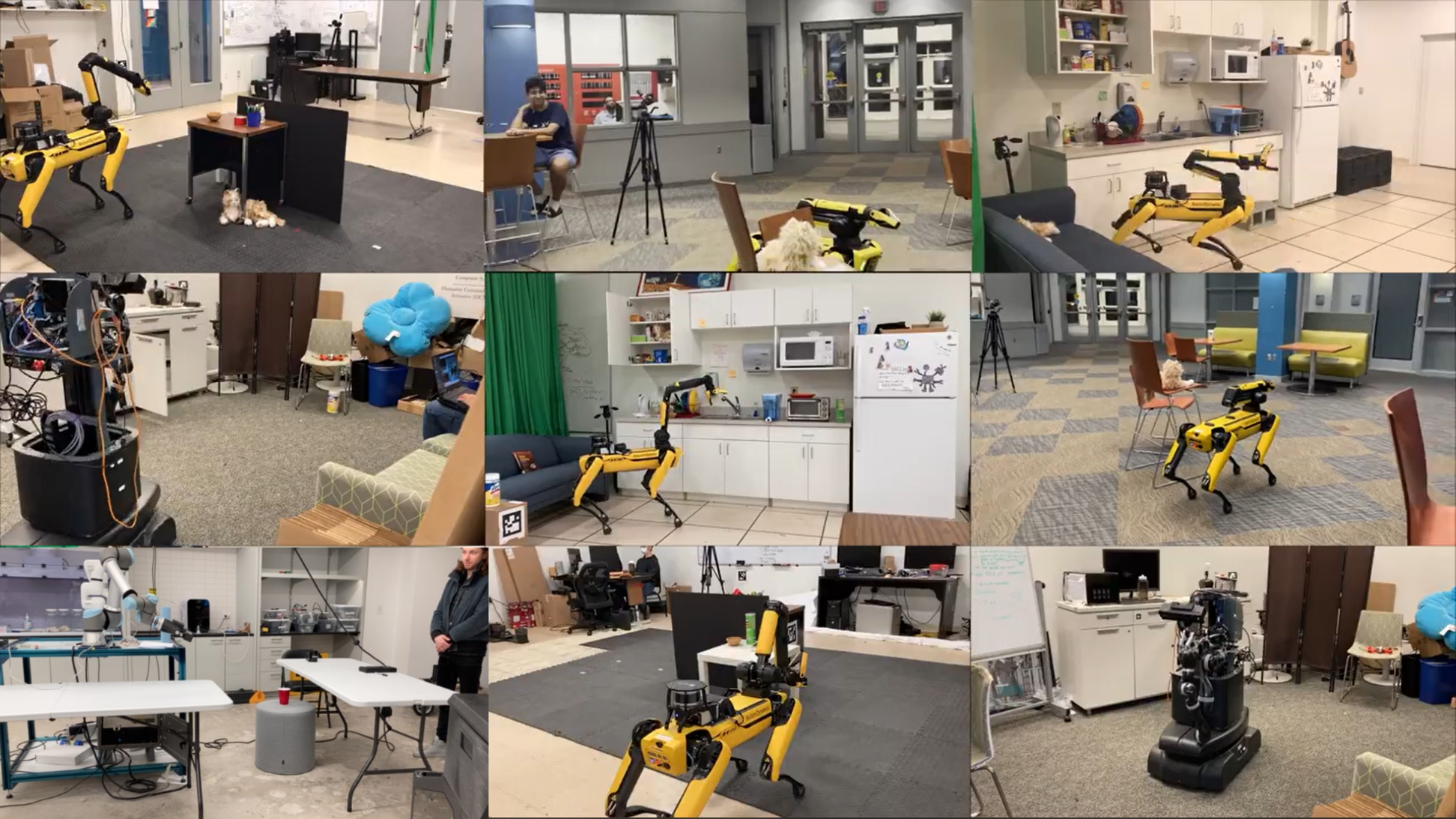}
  \caption{GenMOS enabled different robots to search for objects in different 3D regions. Video: \url{https://youtu.be/TfCe2ZVwypU}
  }
  \vspace{-0.13in}
  \label{fig:demogrid}
\end{figure}

To address these challenges, we present GenMOS (Generalized Multi-Object Search), a general-purpose object search system that is robot-independent and environment-agnostic, the main contribution of this paper. GenMOS builds upon
recent work by \citet{zheng2020multi} which introduced octree belief and a theoretical framework for 3D multi-object search.
GenMOS is a server-client construct.
The server hosts a generic POMDP model of the search agent, which contains the agent's belief state and POMDP models.
The server also maintains an octree representation of the search region's occupancy, used to simulate occlusion-aware observations for belief update. The client sends to the server configurations of the POMDP agent, perception data from robot's sensors, and planning requests, and executes the action returned by the server on the robot (Figure~\ref{fig:system}). The perception data includes point cloud observations of the local search region, object detection results, and localization of the robot view pose. We propose to use point cloud observations in three ways:
(1) to simulate occlusion; (2) to inform occupancy and initialize octree belief; (3) to sample a belief-based graph of view positions.
The server may also actively request information (\eg, additional observation about occupancy), which enables our implementation of hierarchical planning in Section~\ref{sec:hier}.

\begin{figure}[t]
  \centering
  \includegraphics[width=\linewidth]{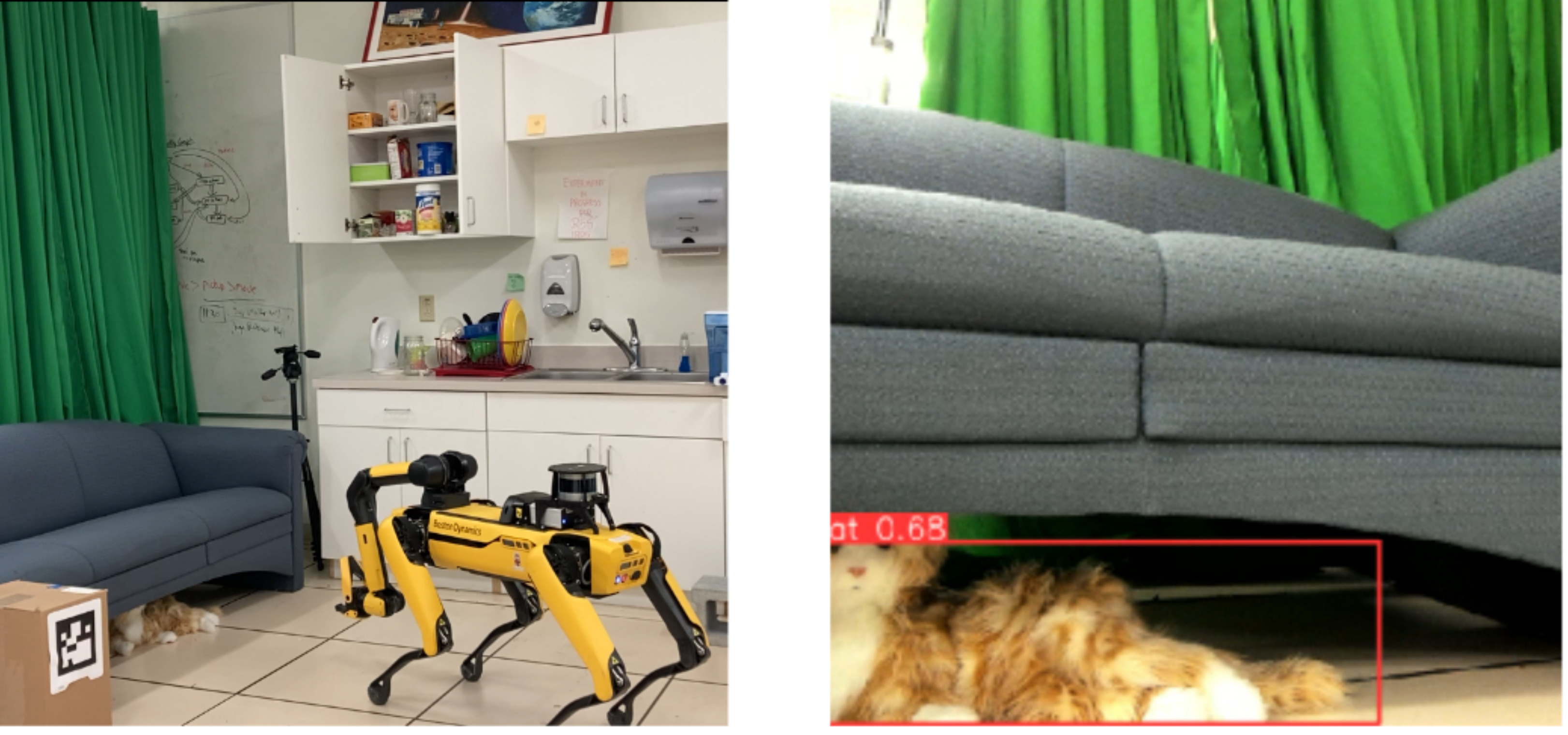}
  \caption{GenMOS enabled a Spot robot to find a toy cat underneath the couch in under one minute. Left: a third-person view of the scene. Right: the RGB image from gripper camera. Note that belief initialization was given only point cloud observations for occupancy and no semantic knowledge (\eg, cats like to hide) was used here.}
  \label{fig:teaser}
  \vspace{-0.2in}
\end{figure}

We implemented GenMOS as a software package leveraging gRPC~\cite{grpc}, a high-performance, cross-platform, and open source framework for remote procedural call (RPC).  We evaluated the package first in simulation, and then deployed it on two robot platforms,\ Boston Dynamics Spot and Kinova MOVO.
\footnote{As a follow-up, we also deployed GenMOS on the Universal Robots UR5e robotic arm (as shown at bottom-left of Figure~\ref{fig:demogrid}); see  \cite{zheng2023generalized} for details.}
In particular, we tasked Spot to search for one or more objects in a region of arranged tables and a kitchen region at the resolution of 0.001m$^3$. Our system enabled it to find, for example, a cat underneath a couch in under one minute (Figure~\ref{fig:teaser}).
Finally, we integrated 3D local search with 2D global search via hierarchical planning to handle larger areas, demonstrating this system in a 25m$^2$ lobby area.

\section{Background}
Recently, \citet{zheng2020multi} introduced a POMDP-based approach for 3D multi-object search (3D-MOS) and a novel octree belief representation for target locations in 3D. Below, we briefly review POMDPs and that approach, including the 3D-MOS model and the octree belief representation.

We emphasize that this paper aims to significantly improve the system-level practicality of the octree-based 3D multi-object search approach from that work, which was only evaluated in an idealistic simulator with actions in cardinal directions and on Kinova MOVO for a proof of concept.


\subsection{POMDP and 3D-MOS}
\label{sec:bg}
A POMDP is defined as a tuple $\langle\Sspace,\Aspace,\Ospace, T,O,$
$R,\gamma\rangle$, where
$\Sspace,\Aspace,\Ospace$ denote the state, action and observation spaces, and  $T,O,R$ are the transition, observation, and reward functions, defined as $T(s,a,s')=\Pr(s'|s,a)$, $O(s',a,o)=\Pr(o|s',a)$,
$R(s,a)\in\mathbb{R}$, respectively.
Since the environment state $s$ is partially observable, the POMDP agent maintains a \emph{belief state} $b_t(s)=\Pr(s|h_t)$ given history $h_t=(ao)_{1:t-1}$. Upon taking action $a_t$ and receiving observation $o_t$, the agent updates its belief by $b_{t+1}(s')=\Pr(s'|h_t,a_t,o_t)=\eta O(s',a_t,o_t)\sum_{s}T(s,a_t,s')b_t(s)$
where $\eta$ is the normalizing constant.
The objective of online POMDP planning is to find a policy $\pi(b_t)\in\Aspace$ which maximizes the
expectation of future discounted rewards ${V^{\pi}}(b_t)=\E\left[\sum_{k=0}^{\infty}\gamma^{k}R(s_{t+k},\pi(b_{t+k}))\ |\ b_t\right]$
with a discount factor
$\gamma$.

A 3D-MOS \cite{zheng2020multi} is an Object-Oriented POMDP (OO-POMDP) \cite{wandzel2019multi}, which is a POMDP with state and observation spaces factored by objects. We refer the reader to the original paper for full details. 3D-MOS is defined as follows:
\begin{itemize}[leftmargin=*,noitemsep]
\item \textbf{State space $\Sspace$.} A state $s=\{s_1,\cdots,s_n,s_r\}$ consists of $n$ target object states and a robot state $s_r$. Each $s_i\in G$ is the 3D location of the target $i$ where $G$ is the discretized \emph{search region}, and $s_r=(q,\mathcal{F})\in\Sspace_r$, where $q=(p,\theta)\in\mathcal{P}\times\SO(3)$ is the 6D camera pose and $\mathcal{F}$ is the set of found objects. The robot state is assumed to be observable.
\item \textbf{Observation space $\Ospace$.} An observation $o$ about the objects, defined as $o=\{(v,d(v) | v\in V)\}$, is a set of labeled voxels within FOV $V$ where a detection function $d(v)$ labels voxel $v$ to be either an object $i\in\{1,\cdots,n\}$, $\textsc{Free}$ or $\textsc{Unknown}$. $\textsc{Free}$ indicates the voxel is a free space or an obstacle, and $\textsc{Unknown}$ indicates occlusion caused by target objects or obstacles in the search region. Refer to \citet{zheng2020multi} for the factorization method.
\item \textbf{Action space $\Aspace$.} Generally, an action can be $\textsc{Move}(s_r,p)$ (move to a reachable position $p\in\mathcal{P}$), $\textsc{Look}(\theta)$ (projects FOV at orientation $\theta\in\SO(3)$), or $\textsc{Find}(i,g)$ (declares object $i$ found at $g\in G$).   In practice, $\textsc{Find}$ is implemented such that targets located within the FOV are marked found.
\item \textbf{Transition function $T$.} Objects are static. $\textsc{Move}(s_r,p)$ and $\textsc{Look}(\theta)$ actions change the robot's camera position and orientation to $p$ and $\theta$ following domain dynamics. $\textsc{Find}(i,g)$ adds $i$ to the set of found objects in the robot state only if $g$ is within the FOV determined by $s_r$.
\item \textbf{Observation function $O$.} The observation model is defined as $\Pr(o_i|s',a)=\Pr(d(s_i')|s',a)$ where $\Pr(d(s_i')=i|s',a)=\alpha$ and $\Pr(d(s_i')=\textsc{Free}|s',a)=\beta$ when $s_i'$ is in $V$. The parameters $\alpha$ and $\beta$ control the reliability of the detector, and are used during octree belief update.
\item \textbf{Reward function $R$.} The reward function is sparse. If $\textsc{Find}$ is taken, yet no new target object is found, the agent receives $\rmin$ (-1000); Otherwise, the agent receives $\rmax$ (+1000). If $\textsc{Move}$ or $\textsc{Look}$ is taken, the agent receives a step cost dependent on the robot state and the action itself.
\end{itemize}

\subsection{Octree Belief}
\label{sec:octree_belief}
An octree belief is a belief state $b_{t}^i$ for object $i$ at time $t$ that consists of an octree and a normalizer.
Denote $\Val_t^i(g^l)$ as the value stored in octree node at $g^l\in G^l$ at resolution level $l$ that covers a cubic volume of $(2^l)^3$. $\Val_t^i(g^l)$ represents the unnormalized probability that object $i$ is present at $g^l$. The normalized belief at $g^l$ is given by $b_t^i(g^l)=\frac{\Val_t^i(g^l)}{\Norm_t}$, where the normalizer $\Norm_t=\sum_{g\in G}\Val_t^i(g)$ equals to the sum of values stored in all nodes at the ground resolution level. The set of nodes at resolution level $k<l$ that reside in a subtree rooted at $g^l$ is denoted $\textsc{Ch}^k(g^l)$.
The value stored in a node equals to the sum of values stored in its children.
 Querying the probability at any node can be done by setting a default value for $\Val_0^i(g)=1$ for all ground cells not yet present in the tree. Then, any node corresponding to $g^l$ has a default value of $\Val_0^i(g^l)=|\textsc{Ch}^0(g^l)|$. Updating the octree belief is exact and efficient with a complexity of $O(|V|\log(|G|))$.
 Sampling is also exact and efficient with a complexity of $O(\log(|G|)$. For full details, please refer to \citet{zheng2020multi}. Here, we define two concepts used later:

 \labelText{\textbf{Definition 1}}{def:3dmos:default_val} \textbf{(Default value).}
  The \emph{default value} of an octree belief node, denoted $\Val_0^i(g^l)$, is the value \emph{before} the node is inserted into the octree.

\labelText{\textbf{Definition 2}}{def:3dmos:initial_val} \textbf{(Initial value).}
  The \emph{initial value} of an octree belief node, denoted $\Val_1^i(g^l)$, is the value \emph{when} the node is inserted into the octree.

\section{Problem Statement}
A robot is tasked to search for one or multiple objects in a 3D region. The robot is assumed to be able to localize itself within the region.
The robot can control the 6-DoF viewpoint (position and orientation) of its camera within a known, continuous reachable space $\mathcal{R}\subseteq \mathcal{P}\times\SO(3)$, where $\mathcal{P}$ is the set of reachable positions. The robot can also receive observations including point cloud and object detection results. The observations are subject to limited FOV, occlusion, noise and errors. To perform object search, the robot should move its camera to different poses (\ie, viewpoints) sequentially to perceive the search region, and eventually declare the target object(s) to be found autonomously. The search performance is evaluated based on the amount of time taken to find objects and the success rate. To be useful for downstream tasks, the robot should identify the 3D locations of found objects.

\section{GenMOS: Generalized Multi-Object Search}
\label{sec:system}
\begin{figure}[t]
  \centering
  \includegraphics[width=\linewidth]{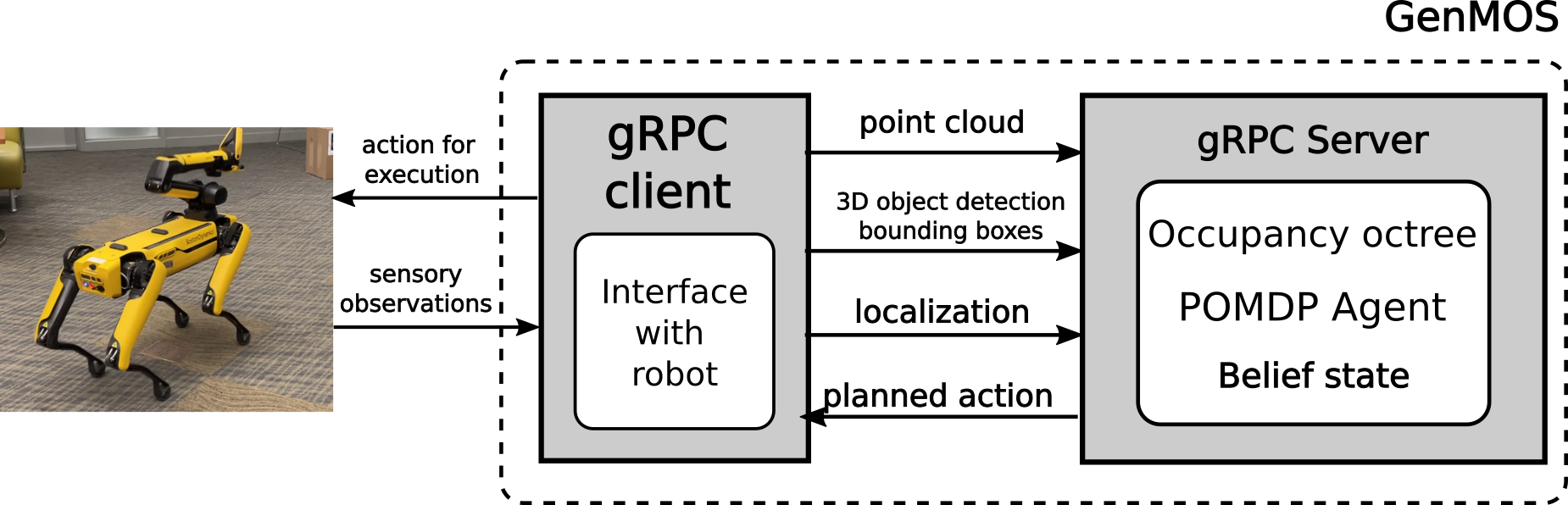}
  \caption{Overview of the GenMOS system. See \ref{sec:system} for description.}
  \label{fig:system}
  \vspace{-0.2in}
\end{figure}

The GenMOS system (Figure~\ref{fig:system}) is a client-server construct. In a nutshell, the server maintains a 3D-MOS agent of the search task and the client acts as a bridge between the 3D-MOS agent and the physical robot in the perception-action loop. Initially, the client requests the \emph{instantiation} of the 3D-MOS agent by providing configurations and a point cloud of the search region. Then, iteratively at each decision step, the client (1) requests an action to be planned by the server, (2) executes the action on the physical robot, and (3) sends to the server object detection results and point cloud from the robot's perception module to update the server's belief state and model of the search region. These steps repeat until all targets are found or the time budget is reached. Below, we detail several aspects of this process.

\begin{figure}[t]
  \centering
  \includegraphics[width=\linewidth]{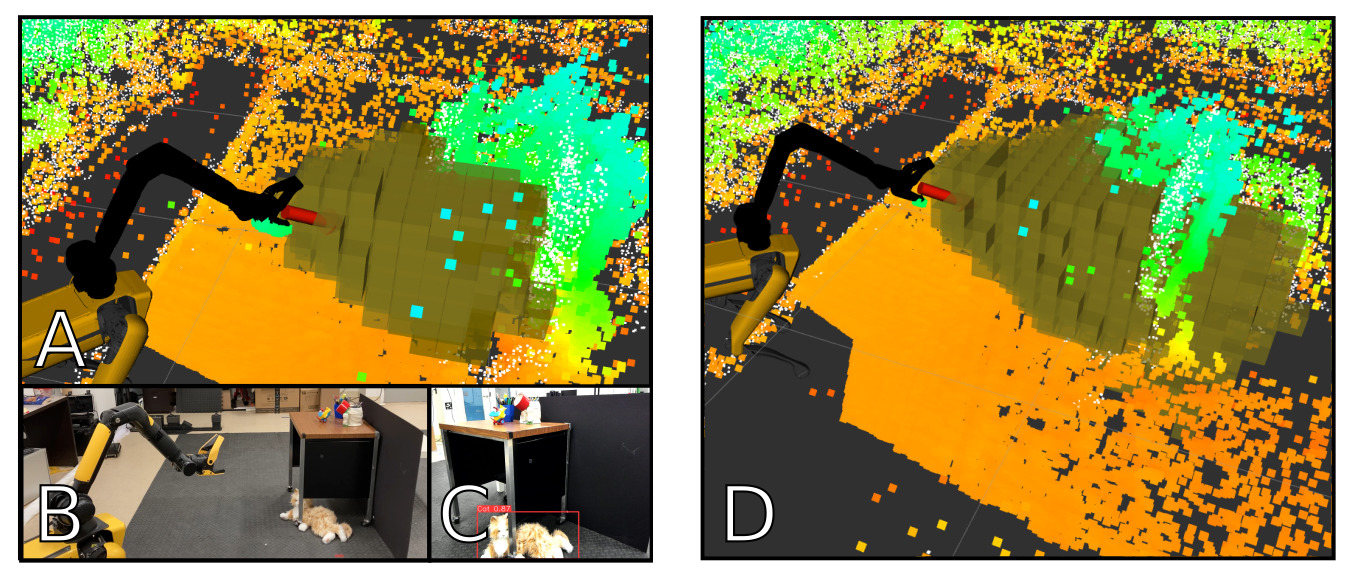}
  \caption{For belief update, GenMOS samples a volumetric observation (a set of labeled voxels within the viewing frustum) that considers occlusion based on the occupancy octree dynamically built from point cloud (A). Not enabling occlusion (D) leads to mistaken invisible locations as free. The robot is looking at a table corner (B) with its view blocked by the table and the board (C).}
  \label{fig:fov}
  \vspace{-0.1in}
\end{figure}

\subsection{Occupancy Octree From Point Cloud}
Internally, the server maintains an octree representation of the search region's occupancy, used to simulate occlusion-aware observations for belief update (Figure.~\ref{fig:fov}). Client sends point cloud observations to the server to initialize and update the server's model of the search region. The server converts the point cloud into an \emph{occupancy octree} (similar to OctoMap \cite{hornung2013octomap}), where a leaf node in the tree has an associated value of occupancy (0 for free, and 1 for occupied). Using this occupancy octree, the server samples volumetric observations during belief update via ray tracing, where occupied nodes block the FOV and cause occlusion. It is also used for sampling the view positions graph to avoid collision (Section~\ref{sec:vgsample}).

\subsection{Prior Initialization of Octree Belief}
\label{ref:prioroctree}
A basic question when instantiating a POMDP agent is: what should be the initial belief? GenMOS uses octree belief, which covers, by definition, a cubic volume. However, the actual feasible search region is likely not cubic, and often irregular. This issue, not addressed in \citet{zheng2020multi},
may cause the robot to constantly falsely believe that the targets are at infeasible locations, which impacts search behavior.

To address this problem, we propose an efficient algorithm for initializing an octree belief over an arbitrary search region (Algorithm~\ref{alg:belief_init}). Recall that $G$ denotes the entire 3D grid map at ground resolution level underlying an octree belief. Suppose $G_*\subseteq G$ is the subset of grids in $G$ that make up the search region.
The high-level idea is as follows:
(1) First, set the \emph{default value} of all ground-level nodes in the octree belief to 0. (2) Then, through a sample-based procedure (with $N$ samples), ground-level nodes whose 3D positions lie within the given search region $G_*$ have their default values changed to 1. This effectively reduces the sample space of the octree belief to be within the search region $G_*$.
Additionally, if we are given a prior distribution,  $\PriorVal^i:G_*^l\rightarrow\mathbb{R}$, we can initialize the octree belief accordingly: during (2), if $\PriorVal^i(g^l)$ is defined at octree belief node $g^l\in G_*^l$, and $g^l$ is the parent (or self) of some ground level node $g\in G$ sampled during step (2), then $\Val_1^i(g^l)$, the \emph{initial value} at $g^l$ is set as $\Val_1^i(g^l)\gets\PriorVal^i(g^l)$.
This algorithm has a complexity of $O(N(\log(|G|))^2)$.

In practice, the server determines the search region $G_*$ based on the occupancy octree constructed from point cloud observations. In our experiments, we assign a prior value of $100\times ((2^k)^3)$ to occupied nodes in the octree at the resolution level $k=2$, and we set the number of samples $N=3000$.



\begin{algorithm}
\caption{OctreeBeliefInit {\small $(m, G_*, \PriorVal^i)\rightarrow b_1^i$}}
\label{alg:belief_init}
\LinesNumbered
\SetNoFillComment
\SetKwInOut{Input}{input}
\SetKwInOut{Output}{output}
\SetKwInOut{Parameter}{param}
\Input{$m$: octree dimension ($|G|=m^3$); $G_*$: the (potentially irregular) search region ($G_*\subseteq G$); $\PriorVal^i$: $G_*^l\rightarrow\mathbb{R}$ (map of prior values).}
\Parameter{$N$: number of samples; $B$: a 3D box, satisfying $G_*\subseteq B \subseteq G$.}
\Output{$b_1^i$: the initialized octree belief.}
Initialize octree $\Psi(b_0^i)$; $\forall g\in G$, set $\Val_0^i(g)=0$\;
\For{$i\in \{1,\cdots,N\}$}{
  Set $l=0$; Sample $g^l\sim B$ \tcp*{sample at ground level}
  \While{$l\leq \log_2m$}{
    \If{$g^l\in G_*^l$}{
      Add $g^l$ to $\Psi(b_1^i)$\tcp*{insert $g^l$ to the octree of $b_1^i$}
      $\Val_0^i(g^l)=|\textsc{Ch}^0(g^l)|$\tcp*{reset default value}
      \If{$g^l\in \PriorVal^i$ }{
         {\small $\Val_1^i(g^l)\gets\PriorVal^i(g^l)$}\tcp*{set initial value; otherwise $\Val_1^i(g^l)\gets \Val_0^i(g^l)$}
      }  
      \tcp{ensure parent value is sum of children}
      Update parent values at $g^{l+1}\cdots g^{m}$\;
      \If{$\Val_1^i(g^l)=\Val_0^i(g^l)$}{remove children of $g^l$ \tcp*{pruning}}
    }
    $l \gets l + 1$\;
  }
}
$\Norm_1\gets\Val_1^i(g^m)$ \tcp*{normalizer set to root's value}
\end{algorithm}

\subsection{Sampling Belief-Dependent View Position Graph}
\label{sec:vgsample}
To enable planning over the continuous space of viewpoints $\mathcal{R}\subseteq \mathcal{P}\times\SO(3)$,
GenMOS samples a view position graph $\mathcal{G}_t=(\mathcal{P}_V, \mathcal{E}_M)$ based on the current octree belief and the occupancy octree. Given an occupancy octree, we first sample a set of non-occupied positions $\mathcal{P}_V$ from $\mathcal{P}$ with a minimum separation threshold, (\eg, 0.75m)
and associate with each position a score representing the belief around that position by querying octree belief at lower resolution. Then, we select top-$K$
(\eg, $K=10$)
nodes ranked by their scores and insert edges such that each node has a limited degree. A $\textsc{Move}(s_r,p_v)$ action then moves the robot to a viewpoint position $p_v\in\mathcal{P}_V$ on the graph. In the transition model, a $\textsc{Look}(\phi)$ action is implicitly enforced after a $\textsc{Move}$ action is taken, where $\phi$ is the orientation facing the an unfound object (contained in $s$, input to the transition model). The graph is resampled at time $t+1$ \emph{if} the sum of the belief probabilities covered by all positions in $\mathcal{G}_t$ is below a threshold (\eg, 0.4).

\subsection{Object Detection}
GenMOS considers generic 3D object detection bounding boxes. The box's size plays a role in the octree belief update, as it influences the volumetric observation. When 3D object detection is not available on the robot, the system can also consume label-only detections based on images, which essentially correspond to 3D object detections where all voxels within the FOV are labeled by the detected object.

\subsection{Planning}
When the server receives planning requests from the client, it plans an action using POUCT \cite{silver2010monte}, a sample-based online POMDP planning algorithm based on Monte Carlo Tree Search. A heuristic rollout policy $\pi_{\text{rollout}}(s)\in\Aspace$ ($s\sim b_t$) is used that samples uniformly among $\textsc{Move}$ actions moving closer towards any target, or $\textsc{Find}$ if a target is within FOV.\footnote{The parallel, multi-resolution POUCT (MR-POUCT) method \cite{zheng2020multi} was not attempted here for simplicity; an efficient implementation of MR-POUCT is left for future work. Simulation results in Section~\ref{tab:results} indicate competency of the planning method implemented in our GenMOS package.}

\begin{figure}[t]
  \centering
  \includegraphics[width=\linewidth]{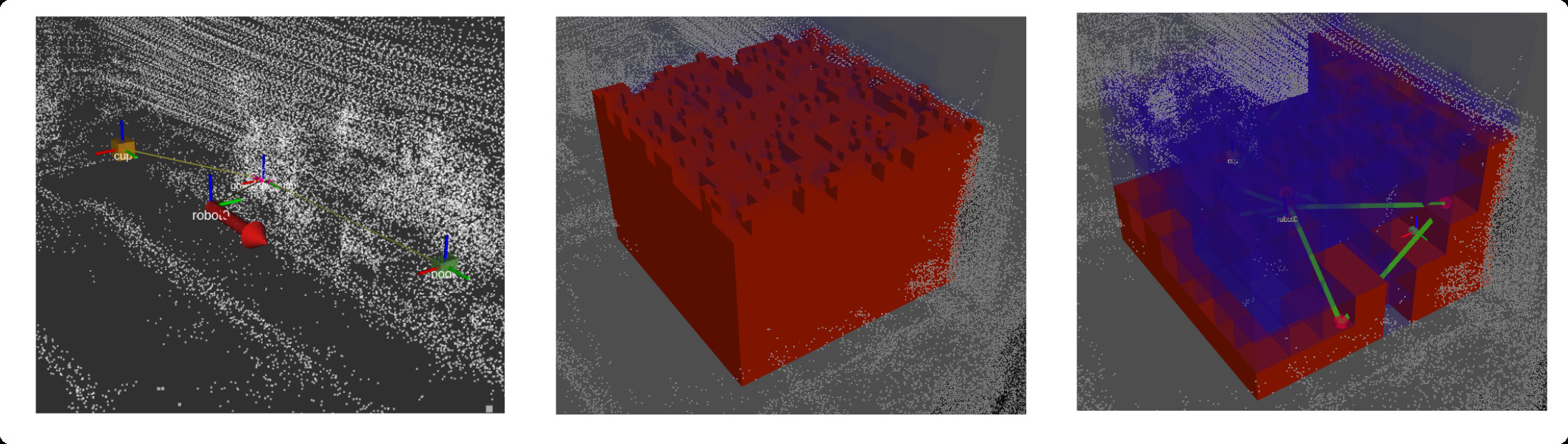}
  \caption{Left: Simulation environment where the pose of the robot's viewpoint is represented by the red arrow, and the two target objects are represented by orange and green cubes. Middle: initialized octree belief given uniform prior; Right: initialized octree belief given occupancy-based prior constructed from point cloud. Colors indicate strength of belief, from red (high) to blue (low).}
  \label{fig:sim_env}
  \vspace{-0.2in}
\end{figure}

\begin{figure*}[htb]
  \centering
\makebox[\textwidth][c]{\includegraphics[width=0.98\textwidth]{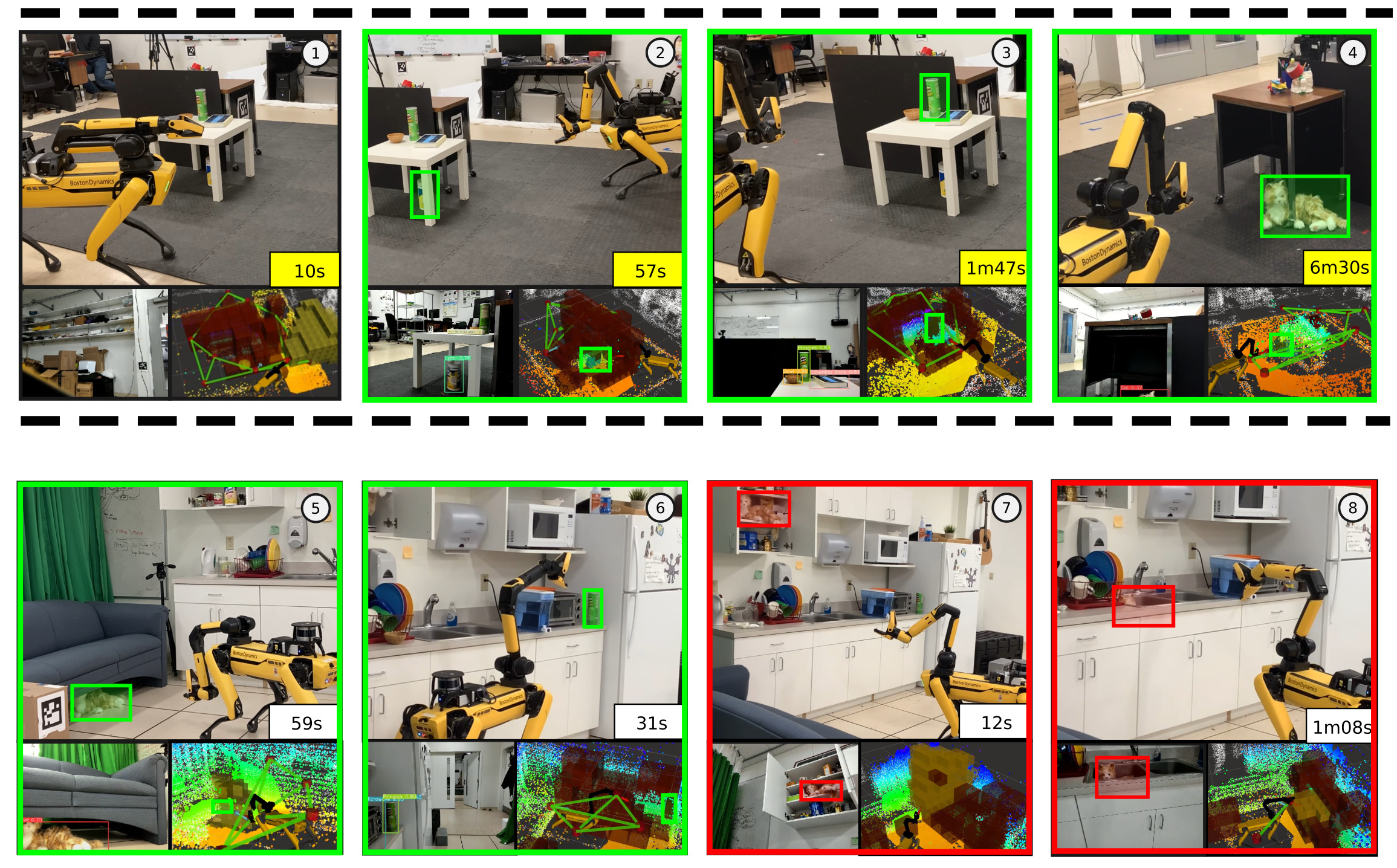}}
  \caption{Key frames from local region search trials on Spot. Each frame consists of three images: a third-person view (top), an image from Spot's gripper camera with object detection (bottom left), and a combined visualization of GenMOS internals (octree belief, viewpoint graph, and local point cloud). Green boxes indicate the marked object is found. Red boxes indicate failure of finding the object due to false negatives in object detection. The yellow or white box in each frame indicates the amount of time passed since the start of the search. Frames at the top row belong to a single trial in the tables region, while frames at the bottom row belong to distinct trials in the kitchen region. The top row (1-4) shows that GenMOS enabled Spot to successfully find multiple objects in the tables region: Lysol under the white table (2), Pringles on the white table (3), and the Cat on the fllor under the wooden table (4). The bottom row shows that GenMOS enabled Spot to find a Cat underneath the couch (5), and the Pringles at the countertop corner (6). (7-8) show a failure mode, where the GenMOS planned a reasonable viewpoint, while the object detector failed to detect the object (Cat) on the shelf or in the sink.}
  \label{fig:spot_examples}
  \vspace{-0.15in}
\end{figure*}

\subsection{Implementation}
We implemented GenMOS as a software package leveraging gRPC~\cite{grpc}.\footnote{ \url{https://github.com/zkytony/genmos_object_search}.} The core (GenMOS server and client) were written in Python, while point cloud processing procedures were in C++. We used \texttt{pomdp-py} \cite{pomdp-py-2020} for POMDP modeling and planning. Using gRPC, our package is independent of, thus integrable to any particular robotic middleware (\eg, ROS \cite{quigley2009ros}, ROS 2 \cite{ros2}, or Viam \cite{viam}). Refer to the appendix for more details, including the gRPC protocol.

\section{Evaluation}
We test two hypotheses through our evaluation: (1) The approach taken by GenMOS based on 3D-MOS, octree belief, and view position graph is effective for 3D object search; (2) The package does enable real robots to search for and find objects in 3D regions in different environments within a reasonable time budget.

To test the first hypothesis, we conduct an experiment in simulation (Section~\ref{sec:sim}).
To test the second hypothesis, we deploy our package on two robot platforms, Boston Dynamics Spot and Kinova MOVO, searching in different local regions, and we also implement a preliminary hierarchical planning algorithm for a demonstration of Spot searching over a larger lobby area (Section~\ref{sec:hier}).


\subsection{Simulation}
\label{sec:sim}
We tasked a simulated robot (represented as an arrow for its viewpoint)
to search for two virtual objects (cubes) with volume 0.002m$^3$ each uniformly randomly placed in a region of size 10.2m$^2\times$ 2.4m. The robot's frustum camera model had a FOV angle of 60 degrees, minimum range of 0.2m and maximum range of 2.0m.

We experimented with three types of priors, groundtruth, uniform, and occupancy-based prior, at two different resolution levels, 0.001$m^3$ (octree size 32$\times$32$\times$32) and 0.008$m^3$ (octree size 16$\times$16$\times$16) representing search granularity.  For the best-performing setting (non-groundtruth), we also compared the use of the POUCT planner against two baselines: Random moves to a uniformly sampled view position graph node (Section~\ref{sec:vgsample}), and Greedy is a next-best view planner that moves to the view position graph node that is closest to the highest belief location for some target. Both baseline planners take $\textsc{Find}$ upon target detection.

We evaluated the search performance by four metrics: total path length traversed during search (Length), total planning time (Planning time), total system time  (Total time), and success rate. Total system time included time for planning, executing navigation actions, receiving observations, belief update and visualization; the simulated robot's translational velocity was 1.0m/s, and its rotational velocity was 0.87rad/s.

\begin{table}[t]
  \centering
    \centering
  \renewcommand{\arraystretch}{1.2}{
    \resizebox{\linewidth}{!}{%
  \begin{tabular}{lllll}
        \toprule
        Prior type (resolution)     & Length     & Planning  & Total      & success \\
        \multicolumn{1}{r}{with POUCT}  & (m)        & time (s)  & time (s)   & rate \\
       \midrule
        Uniform (0.008m$^3$)   &  22.13    & 24.28          & 166.18        & 50\%\\
        Occupancy (0.008m$^3$) &  23.89    & 22.66       & 159.10        & 60\%\\
        Uniform (0.001m$^3$)   &  6.42     & 10.47          & 99.66         & 90\%\\
    Occupancy (0.001m$^3$)$^*$ &  3.22    & 7.42            & 64.12         & 100\%\\
    Groundtruth            &  0.44     & 1.97           & 17.82         & 100\%\\
    \midrule
    \multicolumn{1}{r}{ $^*$with Random}  & 12.18  & 0.19 & 167.20 & 55\%\\
    \multicolumn{1}{r}{ $^*$with Greedy}  & 3.48  & 0.12 & 81.80 & 85\%\\
        \bottomrule
  \end{tabular}
  }}
  \caption{Simulation results. We compare the search performance between different prior
    belief and resolution settings. The results for the first three colums are averaged over 20 trials. }
  \label{tab:results}
  \vspace{-0.2in}
\end{table}

We performed 20 search trials per setting and report the average of each metric in Table~\ref{tab:results}.  Each trial was allowed 180s total system time (excluding the time for visualization). Experiments were run on a computer with i7-8700 CPU. Results indicate that the system achieved high success rate especially at high resolution under occupancy-based prior. We observed that searching with a resolution level more coarse than the target size hurts performance, while having occupancy-based prior improves. Additionally, Greedy was much faster than POUCT in planning time yet lead to lower success rate within the time budget and longer total time than using POUCT.  Our intuition is that, while Greedy prioritizes looking at a location with the highest belief, POUCT considers the search of multiple objects in a sequence.



\subsection{Evaluation on Real Robots}
\label{sec:spot}

\textbf{Boston Dynamics Spot.} We first deployed our system to Spot \cite{bdspot} with a robotic arm by writing a client for GenMOS
that interfaces with the Spot SDK.\footnote{We integrated Spot SDK with
  ROS \cite{quigley2009ros} to use RViZ \cite{kam2015rviz};
  Our computer that ran GenMOS for Spot has an i7-9750H CPU with an RTX 2060 GPU.}
We tasked the robot to search in two local regions in different rooms (Figure~\ref{fig:spot_examples}). The first region (of size 9m$^2\times 1.5$m) arranged two tables and a separation board which created occlusion. The second region (of size 7.5m$^2\times 2.2$m) was a kitchen area with a countertop, a shelf, and a couch. In both, the resolution of the octree belief was set to 0.001$m^3$ with a size of 32$\times$32$\times$32. Occupany-based prior from point cloud was used. The robot was given at most 10 minutes to search. We collected a dataset of 230 images and trained a YOLOv5 detector \cite{glenn_jocher_2020_4154370} with 1.9 million parameters for the objects of interest.
2D bounding boxes were projected to 3D using depth from gripper camera.

\begin{figure}[t]
  \centering
  \includegraphics[width=\linewidth]{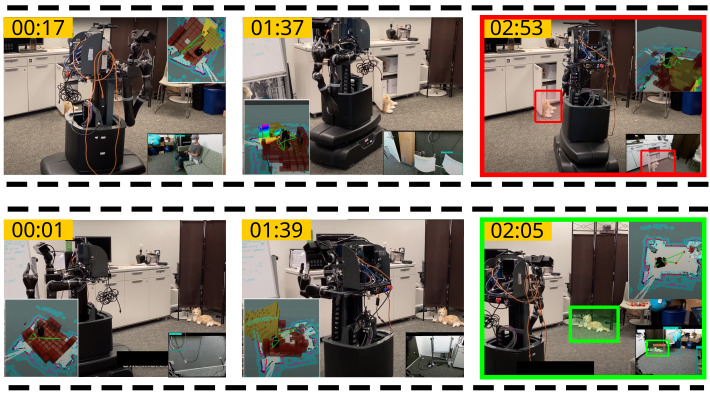}
  \caption{Key frames from trials on MOVO. Top: the Cat was lying next to the opened door. MOVO looked in the right direction but the detector missed it. Bottom: Although the detector missed at first, MOVO recovered and found the target eventually. Image from Kinect and GenMOS internal visualization are shown in each frame. }
  \label{fig:movo_examples}
  \vspace{-0.15in}
\end{figure}

\begin{figure}[t]
  \centering
  \includegraphics[width=\linewidth]{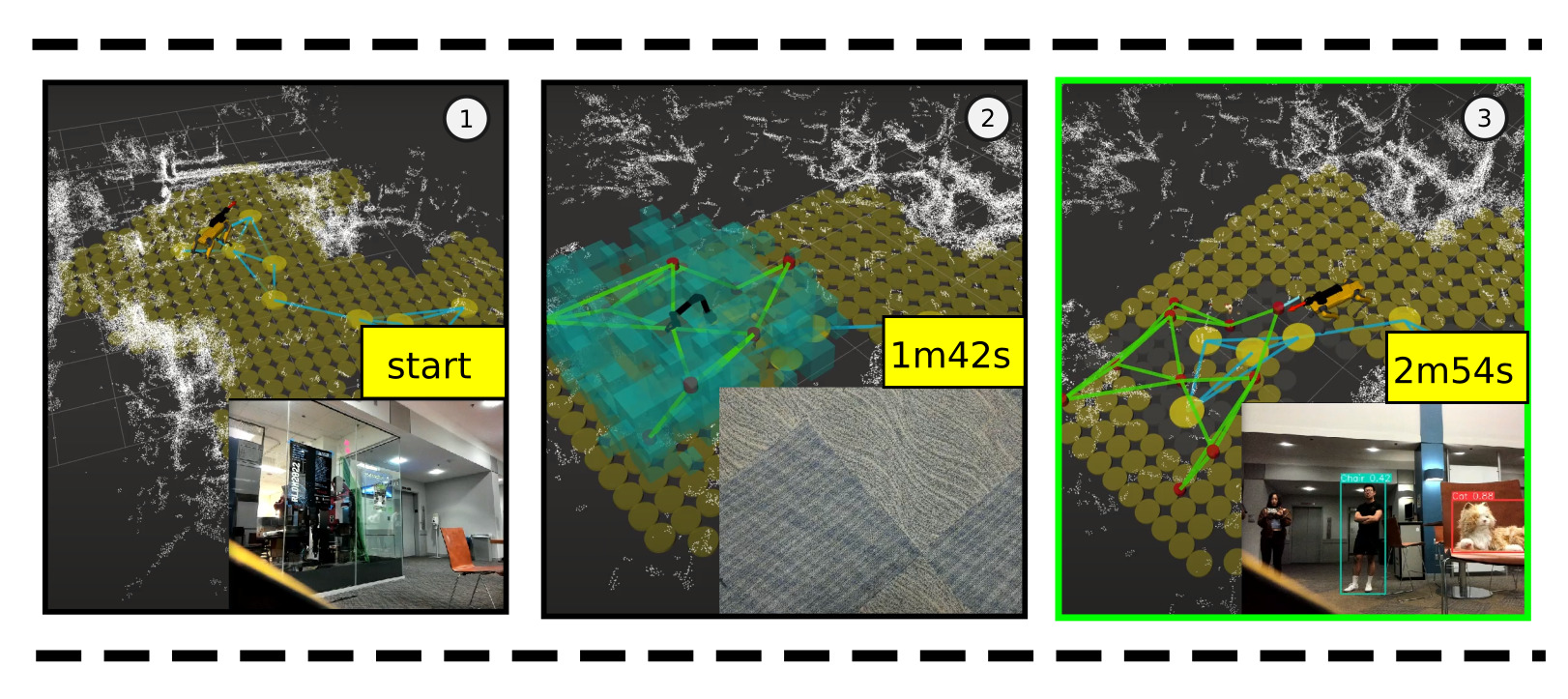}
  \caption{Demonstration of hierarchical planning where a 2D global search was integrated
    with 3D local search through the \emph{stay} action \cite{zheng2022towards}. This system enabled the Spot robot to find a Cat in a lobby area within 3~minutes. (1)~Initial state; (2)~searching in a 3D local region; (3)~the robot detected the Cat and the search finished.}
  \label{fig:lobby}
    \vspace{-0.2in}
\end{figure}

\textbf{Kinova MOVO.} We then deployed GenMOS to Kinova MOVO (Figure~\ref{fig:movo_examples}), which has a mobile base, an extensible torso, and a head that can pan and tilt, equipped with a Kinect V2 RGBD camera.  Similar to Spot, we deployed
GenMOS to MOVO by integrating the GenMOS gRPC client with the perception,
navigation and control stacks of MOVO, which was based on ROS Kinetic.
We evaluated the resulting object search
system in a small living room environment (of size 10.5m$^2\times$1.5m) searching for a Cat.

\textbf{Results.} Figure~\ref{fig:spot_examples} and Figure~\ref{fig:movo_examples} contain illustrations of key frames during the search trials with Spot and MOVO, respectively. Video footages of the search together with belief state visualization are available in the supplementary video. In the arranged tables region, our system enabled Spot to simultaneously search for four objects (Cat, Pringles, Lysol, and ToyPlane), and successfully found three objects in 6.5 minutes. In the kitchen region, our system enabled Spot to find a Cat placed underneath a couch within one minute. Compared to Spot, however, MOVO was less agile and prone to collision while navigating between viewpoints during the search. Nevertheless, GenMOS enabled MOVO to perform search and found Cat on the floor in around 2 minutes.

We do observe that search success was impacted by false negatives from the object detector, as well as conservative viewpoint sampling for obstacle avoidance. The latter prevented the robot to plan top-down views from \eg, directly above the countertop. Overall, our system enabled different robots to search for objects in different environments within a moderate time budget.

\subsection{Extension to Hierarchical Planning}
\label{sec:hier}
We envision the integration of our 3D local search algorithm with a global search algorithm so that a larger search space can be handled. To this end, we implemented a hierarchical planning algorithm that contains a 2D global planner (with the same multi-object search POMDP model in Section \ref{sec:bg} but in 2D), where the global planner had a \emph{stay} action (no viewpoint change) which triggers
the initialization of a 3D local search agent. In particular, in our implementation, when the planner \emph{decided} to search locally, we let the server send a message that triggered the client to send over an update search region request to initialize the local 3D search agent.


The starting belief of the 3D local agent was initialized based on the 2D global belief; the 2D global belief was in turn updated by projecting the 3D field of view down to 2D. We set the resolution of 2D search to be 0.09$m^2$, and the resolution of 3D search to be 0.001$m^3$. We tested this system in a lobby area of size 25m$^2\times 1.5$m, where the robot was tasked to find the toy cat on a tall chair (Figure~\ref{fig:lobby}). The search succeeded within three minutes, covering roughly 15m$^2$.

\section{Related Work}
Beginning with CARMEN~\citep{montemerlo2003perspectives}, open source libraries for
SLAM have greatly lowered the barrier to entry into
robotics~\citep{grisetti2007improved, hess2016real}.  Similarly, for
motion planning, libraries such as
OMPL~\citep{sucan2012the-open-motion-planning-library} and
MoveIt!~\citep{chitta2016moveit} have broadened access to motion
planning to a variety of different robotic platforms.  Our work aims
to do the same thing for object search.

\citet{wixson1994using} remarks that selecting views for object search in a local region is a harder problem than the selection of which region to search in. Most works that demonstrate robotic search within a search region reduce the problem to 2D \cite{aydemir2013avo,wandzel2019multi,li2016act,zeng2020semantic,bejjani2021occlusion,holzherr2021efficient,giuliari2021pomp++,schmalstieg2022learning}. For a literature survey and taxonomy of object search, refer to \cite{zheng2023generalized}.

Deep learning methods that typically map raw observations to actions \cite{yang2018visual,chaplot2020object,mayo2021visual,deitke2022procthor,schmalstieg2022learning} can enable 3D object search, yet it is hard to train such a model on a robot and ensure generalization to a new real-world environment; ongoing work (\eg, \cite{schmalstieg2022learning}) is addressing this challenge.
In contrast, GenMOS only requires basic perception capabilities such as object detection and localization to enable object search.
Our work extends the 3D-MOS \cite{zheng2020multi} approach for 3D local region search, by proposing algorithms for making octree belief more applicable, and by developing a practical system that uses octree belief as the representation of uncertainty.

\section{Conclusion and Future Work}
We introduced GenMOS, the first robot-independent and environment-agnostic system of multi-object search in 3D regions. We implemented the system as a package and evaluated it in simulation and on two real robot platforms, and demonstrated an extension for searching over a larger area. Future work should investigate the integration of common sense or correlation \cite{zheng2022towards} and spatial language \cite{sloop-roman-2020}, and search during map exploration or involving manipulation.

\bibliography{root}

\begin{thebibliography}{40}
\providecommand{\natexlab}[1]{#1}
\providecommand{\url}[1]{#1}
\csname url@samestyle\endcsname
\providecommand{\newblock}{\relax}
\providecommand{\bibinfo}[2]{#2}
\providecommand{\BIBentrySTDinterwordspacing}{\spaceskip=0pt\relax}
\providecommand{\BIBentryALTinterwordstretchfactor}{4}
\providecommand{\BIBentryALTinterwordspacing}{\spaceskip=\fontdimen2\font plus
\BIBentryALTinterwordstretchfactor\fontdimen3\font minus
  \fontdimen4\font\relax}
\providecommand{\BIBforeignlanguage}[2]{{%
\expandafter\ifx\csname l@#1\endcsname\relax
\typeout{** WARNING: IEEEtranN.bst: No hyphenation pattern has been}%
\typeout{** loaded for the language `#1'. Using the pattern for}%
\typeout{** the default language instead.}%
\else
\language=\csname l@#1\endcsname
\fi
#2}}
\providecommand{\BIBdecl}{\relax}
\BIBdecl

\bibitem[Nourbakhsh et~al.(2005)Nourbakhsh, Sycara, Koes, Yong, Lewis, and
  Burion]{nourbakhsh2005human}
I.~R. Nourbakhsh, K.~Sycara, M.~Koes, M.~Yong, M.~Lewis, and S.~Burion,
  ``Human-robot teaming for search and rescue,'' \emph{IEEE Pervasive
  Computing}, vol.~4, no.~1, pp. 72--79, 2005.

\bibitem[Kamegawa et~al.(2020)Kamegawa, Akiyama, Sakai, Fujii, Une, Ou,
  Matsumura, Kishutani, Nose, Yoshizaki, et~al.]{kamegawa2020development}
T.~Kamegawa, T.~Akiyama, S.~Sakai, K.~Fujii, K.~Une, E.~Ou, Y.~Matsumura,
  T.~Kishutani, E.~Nose, Y.~Yoshizaki \emph{et~al.}, ``Development of a
  separable search-and-rescue robot composed of a mobile robot and a snake
  robot,'' \emph{Advanced Robotics}, vol.~34, no.~2, pp. 132--139, 2020.

\bibitem[Sharma et~al.(2021)Sharma, Torralba, and Andreas]{sharma2021skill}
P.~Sharma, A.~Torralba, and J.~Andreas, ``Skill induction and planning with
  latent language,'' \emph{arXiv preprint arXiv:2110.01517}, 2021.

\bibitem[Ahn et~al.(2022)Ahn, Brohan, Brown, Chebotar, Cortes, David, Finn, Fu,
  Gopalakrishnan, Hausman, Herzog, Ho, Hsu, Ibarz, Ichter, Irpan, Jang, Ruano,
  Jeffrey, Jesmonth, Joshi, Julian, Kalashnikov, Kuang, Lee, Levine, Lu, Luu,
  Parada, Pastor, Quiambao, Rao, Rettinghouse, Reyes, Sermanet, Sievers, Tan,
  Toshev, Vanhoucke, Xia, Xiao, Xu, Xu, Yan, and Zeng]{saycan2022arxiv}
M.~Ahn, A.~Brohan, N.~Brown, Y.~Chebotar, O.~Cortes, B.~David, C.~Finn, C.~Fu,
  K.~Gopalakrishnan, K.~Hausman, A.~Herzog, D.~Ho, J.~Hsu, J.~Ibarz, B.~Ichter,
  A.~Irpan, E.~Jang, R.~J. Ruano, K.~Jeffrey, S.~Jesmonth, N.~Joshi, R.~Julian,
  D.~Kalashnikov, Y.~Kuang, K.-H. Lee, S.~Levine, Y.~Lu, L.~Luu, C.~Parada,
  P.~Pastor, J.~Quiambao, K.~Rao, J.~Rettinghouse, D.~Reyes, P.~Sermanet,
  N.~Sievers, C.~Tan, A.~Toshev, V.~Vanhoucke, F.~Xia, T.~Xiao, P.~Xu, S.~Xu,
  M.~Yan, and A.~Zeng, ``Do as i can and not as i say: Grounding language in
  robotic affordances,'' in \emph{arXiv preprint arXiv:2204.01691}, 2022.

\bibitem[Kaelbling et~al.(1998)Kaelbling, Littman, and
  Cassandra]{kaelbling1998planning}
L.~P. Kaelbling, M.~L. Littman, and A.~R. Cassandra, ``Planning and acting in
  partially observable stochastic domains,'' \emph{Artificial intelligence},
  vol. 101, no. 1-2, pp. 99--134, 1998.

\bibitem[Wandzel et~al.(2019)Wandzel, Oh, Fishman, Kumar, and
  Tellex]{wandzel2019multi}
A.~Wandzel, Y.~Oh, M.~Fishman, N.~Kumar, and S.~Tellex, ``Multi-object search
  using object-oriented {POMDPs},'' in \emph{2019 International Conference on
  Robotics and Automation (ICRA)}.\hskip 1em plus 0.5em minus 0.4em\relax IEEE,
  2019.

\bibitem[Zeng et~al.(2020)Zeng, R{\"o}fer, and Jenkins]{zeng2020semantic}
Z.~Zeng, A.~R{\"o}fer, and O.~C. Jenkins, ``Semantic linking maps for active
  visual object search,'' in \emph{2020 IEEE International Conference on
  Robotics and Automation (ICRA)}.\hskip 1em plus 0.5em minus 0.4em\relax IEEE,
  2020, pp. 1984--1990.

\bibitem[Zhu et~al.(2017)Zhu, Mottaghi, Kolve, Lim, Gupta, Fei-Fei, and
  Farhadi]{zhu2017target}
Y.~Zhu, R.~Mottaghi, E.~Kolve, J.~J. Lim, A.~Gupta, L.~Fei-Fei, and A.~Farhadi,
  ``Target-driven visual navigation in indoor scenes using deep reinforcement
  learning,'' in \emph{2017 IEEE international conference on robotics and
  automation (ICRA)}.\hskip 1em plus 0.5em minus 0.4em\relax IEEE, 2017, pp.
  3357--3364.

\bibitem[Batra et~al.(2020)Batra, Gokaslan, Kembhavi, Maksymets, Mottaghi,
  Savva, Toshev, and Wijmans]{batra2020objectnav}
D.~Batra, A.~Gokaslan, A.~Kembhavi, O.~Maksymets, R.~Mottaghi, M.~Savva,
  A.~Toshev, and E.~Wijmans, ``Objectnav revisited: On evaluation of embodied
  agents navigating to objects,'' \emph{arXiv preprint arXiv:2006.13171}, 2020.

\bibitem[Zheng et~al.(2021{\natexlab{a}})Zheng, Sung, Konidaris, and
  Tellex]{zheng2020multi}
K.~Zheng, Y.~Sung, G.~Konidaris, and S.~Tellex, ``Multi-resolution {POMDP}
  planning for multi-object search in {3D},'' in \emph{IEEE/RSJ International
  Conference on Intelligent Robots and Systems (IROS)}, 2021.

\bibitem[grp()]{grpc}
``g{PRC} documentation,'' \url{https://grpc.io/docs/}, accessed: 2022.

\bibitem[Zheng(2023)]{zheng2023generalized}
K.~Zheng, ``Generalized object search,'' Ph.D. dissertation, Brown University,
  February 2023.

\bibitem[Hornung et~al.(2013)Hornung, Wurm, Bennewitz, Stachniss, and
  Burgard]{hornung2013octomap}
A.~Hornung, K.~M. Wurm, M.~Bennewitz, C.~Stachniss, and W.~Burgard, ``Octomap:
  An efficient probabilistic 3d mapping framework based on octrees,''
  \emph{Autonomous robots}, vol.~34, no.~3, pp. 189--206, 2013.

\bibitem[Silver and Veness(2010)]{silver2010monte}
D.~Silver and J.~Veness, ``Monte-carlo planning in large {POMDPs},'' in
  \emph{Neural Information Processing Systems}, 2010.

\bibitem[Zheng and Tellex(2020)]{pomdp-py-2020}
K.~Zheng and S.~Tellex, ``pomdp\_py: A framework to build and solve {POMDP}
  problems,'' in \emph{ICAPS 2020 Workshop on Planning and Robotics (PlanRob)},
  2020.

\bibitem[Quigley et~al.(2009)Quigley, Conley, Gerkey, Faust, Foote, Leibs,
  Wheeler, and Ng]{quigley2009ros}
M.~Quigley, K.~Conley, B.~Gerkey, J.~Faust, T.~Foote, J.~Leibs, R.~Wheeler, and
  A.~Y. Ng, ``Ros: an open-source robot operating system,'' in \emph{ICRA
  workshop on open source software}, vol.~3, no. 3.2.\hskip 1em plus 0.5em
  minus 0.4em\relax Kobe, Japan, 2009, p.~5.

\bibitem[Macenski et~al.(2022)Macenski, Foote, Gerkey, Lalancette, and
  Woodall]{ros2}
\BIBentryALTinterwordspacing
S.~Macenski, T.~Foote, B.~Gerkey, C.~Lalancette, and W.~Woodall, ``Robot
  operating system 2: Design, architecture, and uses in the wild,''
  \emph{Science Robotics}, vol.~7, no.~66, p. eabm6074, 2022. [Online].
  Available: \url{https://www.science.org/doi/abs/10.1126/scirobotics.abm6074}
\BIBentrySTDinterwordspacing

\bibitem[via(2022)]{viam}
``{Viam, Inc.}'' \url{https://www.viam.com/}, 2022, accessed: Feb.~2023.

\bibitem[bds()]{bdspot}
``{Boston Dynamics Spot},'' \url{https://www.bostondynamics.com/products/spot},
  accessed: 2019.

\bibitem[Kam et~al.(2015)Kam, Lee, Park, and Kim]{kam2015rviz}
H.~R. Kam, S.-H. Lee, T.~Park, and C.-H. Kim, ``Rviz: a toolkit for real domain
  data visualization,'' \emph{Telecommunication Systems}, vol.~60, no.~2, pp.
  337--345, 2015.

\bibitem[Jocher et~al.(2020)Jocher, Stoken, Borovec, NanoCode012,
  ChristopherSTAN, Changyu, Laughing, tkianai, Hogan, lorenzomammana, yxNONG,
  AlexWang1900, Diaconu, Marc, wanghaoyang0106, ml5ah, Doug, Ingham, Frederik,
  Guilhen, Hatovix, Poznanski, Fang, Yu, changyu98, Wang, Gupta, Akhtar,
  PetrDvoracek, and Rai]{glenn_jocher_2020_4154370}
\BIBentryALTinterwordspacing
G.~Jocher, A.~Stoken, J.~Borovec, NanoCode012, ChristopherSTAN, L.~Changyu,
  Laughing, tkianai, A.~Hogan, lorenzomammana, yxNONG, AlexWang1900,
  L.~Diaconu, Marc, wanghaoyang0106, ml5ah, Doug, F.~Ingham, Frederik, Guilhen,
  Hatovix, J.~Poznanski, J.~Fang, L.~Yu, changyu98, M.~Wang, N.~Gupta,
  O.~Akhtar, PetrDvoracek, and P.~Rai, ``{ultralytics/yolov5: v3.1 - Bug Fixes
  and Performance Improvements},'' Oct. 2020. [Online]. Available:
  \url{https://doi.org/10.5281/zenodo.4154370}
\BIBentrySTDinterwordspacing

\bibitem[Zheng et~al.(2022)Zheng, Chitnis, Sung, Konidaris, and
  Tellex]{zheng2022towards}
K.~Zheng, R.~Chitnis, Y.~Sung, G.~Konidaris, and S.~Tellex, ``Towards optimal
  correlational object search,'' in \emph{IEEE International Conference on
  Robotics and Automation (ICRA)}, 2022.

\bibitem[Montemerlo et~al.(2003)Montemerlo, Roy, and
  Thrun]{montemerlo2003perspectives}
M.~Montemerlo, N.~Roy, and S.~Thrun, ``Perspectives on standardization in
  mobile robot programming: The carnegie mellon navigation (carmen) toolkit,''
  in \emph{Proceedings 2003 IEEE/RSJ International Conference on Intelligent
  Robots and Systems (IROS 2003)(Cat. No. 03CH37453)}, vol.~3.\hskip 1em plus
  0.5em minus 0.4em\relax IEEE, 2003, pp. 2436--2441.

\bibitem[Grisetti et~al.(2007)Grisetti, Stachniss, and
  Burgard]{grisetti2007improved}
G.~Grisetti, C.~Stachniss, and W.~Burgard, ``Improved techniques for grid
  mapping with rao-blackwellized particle filters,'' \emph{IEEE transactions on
  Robotics}, vol.~23, no.~1, pp. 34--46, 2007.

\bibitem[Hess et~al.(2016)Hess, Kohler, Rapp, and Andor]{hess2016real}
W.~Hess, D.~Kohler, H.~Rapp, and D.~Andor, ``Real-time loop closure in 2d lidar
  slam,'' in \emph{2016 IEEE international conference on robotics and
  automation (ICRA)}.\hskip 1em plus 0.5em minus 0.4em\relax IEEE, 2016, pp.
  1271--1278.

\bibitem[{\c{S}}ucan et~al.(2012){\c{S}}ucan, Moll, and
  Kavraki]{sucan2012the-open-motion-planning-library}
I.~A. {\c{S}}ucan, M.~Moll, and L.~E. Kavraki, ``The {O}pen {M}otion {P}lanning
  {L}ibrary,'' \emph{{IEEE} Robotics \& Automation Magazine}, vol.~19, no.~4,
  pp. 72--82, December 2012, \url{https://ompl.kavrakilab.org}.

\bibitem[Chitta(2016)]{chitta2016moveit}
S.~Chitta, ``Moveit!: an introduction,'' in \emph{Robot Operating System
  (ROS)}.\hskip 1em plus 0.5em minus 0.4em\relax Springer, 2016, pp. 3--27.

\bibitem[Wixson and Ballard(1994)]{wixson1994using}
L.~E. Wixson and D.~H. Ballard, ``Using intermediate objects to improve the
  efficiency of visual search,'' \emph{International Journal of Computer
  Vision}, vol.~12, no. 2-3, pp. 209--230, 1994.

\bibitem[Aydemir et~al.(2013)Aydemir, Pronobis, G{\"o}belbecker, and
  Jensfelt]{aydemir2013avo}
A.~Aydemir, A.~Pronobis, M.~G{\"o}belbecker, and P.~Jensfelt, ``Active visual
  object search in unknown environments using uncertain semantics,'' \emph{IEEE
  Transactions on Robotics (T-RO)}, vol.~29, no.~4, pp. 986--1002, Aug. 2013.

\bibitem[Li et~al.(2016)Li, Hsu, and Lee]{li2016act}
J.~K. Li, D.~Hsu, and W.~S. Lee, ``Act to see and see to act: {POMDP} planning
  for objects search in clutter,'' in \emph{2016 IEEE/RSJ International
  Conference on Intelligent Robots and Systems (IROS)}.\hskip 1em plus 0.5em
  minus 0.4em\relax IEEE, 2016.

\bibitem[Bejjani et~al.(2021)Bejjani, Agboh, Dogar, and
  Leonetti]{bejjani2021occlusion}
W.~Bejjani, W.~C. Agboh, M.~R. Dogar, and M.~Leonetti, ``Occlusion-aware search
  for object retrieval in clutter,'' in \emph{2021 IEEE/RSJ International
  Conference on Intelligent Robots and Systems (IROS)}.\hskip 1em plus 0.5em
  minus 0.4em\relax IEEE, 2021, pp. 4678--4685.

\bibitem[Holzherr et~al.(2021)Holzherr, F{\"o}rster, Breyer, Nieto, Siegwart,
  and Chung]{holzherr2021efficient}
L.~Holzherr, J.~F{\"o}rster, M.~Breyer, J.~Nieto, R.~Siegwart, and J.~J. Chung,
  ``Efficient multi-scale {POMDPs} for robotic object search and delivery,'' in
  \emph{2021 IEEE International Conference on Robotics and Automation
  (ICRA)}.\hskip 1em plus 0.5em minus 0.4em\relax IEEE, 2021, pp. 6585--6591.

\bibitem[Giuliari et~al.(2021)Giuliari, Castellini, Berra, Del~Bue, Farinelli,
  Cristani, Setti, and Wang]{giuliari2021pomp++}
F.~Giuliari, A.~Castellini, R.~Berra, A.~Del~Bue, A.~Farinelli, M.~Cristani,
  F.~Setti, and Y.~Wang, ``Pomp++: Pomcp-based active visual search in unknown
  indoor environments,'' in \emph{2021 IEEE/RSJ International Conference on
  Intelligent Robots and Systems (IROS)}.\hskip 1em plus 0.5em minus
  0.4em\relax IEEE, 2021, pp. 1523--1530.

\bibitem[Schmalstieg et~al.(2022)Schmalstieg, Honerkamp, Welschehold, and
  Valada]{schmalstieg2022learning}
F.~Schmalstieg, D.~Honerkamp, T.~Welschehold, and A.~Valada, ``Learning
  long-horizon robot exploration strategies for multi-object search in
  continuous action spaces,'' \emph{arXiv preprint arXiv:2205.11384}, 2022.

\bibitem[Yang et~al.(2019)Yang, Wang, Farhadi, Gupta, and
  Mottaghi]{yang2018visual}
W.~Yang, X.~Wang, A.~Farhadi, A.~Gupta, and R.~Mottaghi, ``Visual semantic
  navigation using scene priors,'' \emph{International Conference on Learning
  Representations (ICLR)}, 2019.

\bibitem[Chaplot et~al.(2020)Chaplot, Gandhi, Gupta, and
  Salakhutdinov]{chaplot2020object}
D.~S. Chaplot, D.~P. Gandhi, A.~Gupta, and R.~R. Salakhutdinov, ``Object goal
  navigation using goal-oriented semantic exploration,'' \emph{Advances in
  Neural Information Processing Systems}, vol.~33, pp. 4247--4258, 2020.

\bibitem[Mayo et~al.(2021)Mayo, Hazan, and Tal]{mayo2021visual}
B.~Mayo, T.~Hazan, and A.~Tal, ``Visual navigation with spatial attention,'' in
  \emph{Proceedings of the IEEE/CVF Conference on Computer Vision and Pattern
  Recognition}, 2021, pp. 16\,898--16\,907.

\bibitem[Deitke et~al.(2022)Deitke, VanderBilt, Herrasti, Weihs, Salvador,
  Ehsani, Han, Kolve, Farhadi, Kembhavi, et~al.]{deitke2022procthor}
M.~Deitke, E.~VanderBilt, A.~Herrasti, L.~Weihs, J.~Salvador, K.~Ehsani,
  W.~Han, E.~Kolve, A.~Farhadi, A.~Kembhavi \emph{et~al.}, ``Procthor:
  Large-scale embodied ai using procedural generation,'' \emph{arXiv preprint
  arXiv:2206.06994}, 2022.

\bibitem[Zheng et~al.(2021{\natexlab{b}})Zheng, Bayazit, Mathew, Pavlick, and
  Tellex]{sloop-roman-2020}
K.~Zheng, D.~Bayazit, R.~Mathew, E.~Pavlick, and S.~Tellex, ``Spatial language
  understanding for object search in partially observed cityscale
  environments,'' in \emph{International Conference on Robot and Human
  Interactive Communication (RO-MAN)}.\hskip 1em plus 0.5em minus 0.4em\relax
  IEEE, 2021.

\bibitem[Varda()]{protobuf}
K.~Varda, ``{Protocol Buffers},''
  \url{http://code.google.com/apis/protocolbuffers/}.

\end{thebibliography}

\newpage
\appendix
\section{Appendix}

\subsection{The gRPC Protocol for GenMOS}
\label{sec:grpc}
In the gRPC framework, remote procedural calls (RPCs) are defined as Protocol Buffer messages \cite{protobuf}. In particular, the key RPCs in our package for GenMOS are as follows:
\begin{itemize}[leftmargin=*,noitemsep]
\item \texttt{CreateAgent}: Upon receiving the POMDP agent configurations from the client, the server prepares for agent creation pending the first \texttt{UpdateSearchRegion} call.

\item \texttt{UpdateSearchRegion}: The client sends over a point cloud of the local search region, and the server creates or updates the occupancy octree about the search region.

\item \texttt{ProcessObservation}: The client requests belief update by sending observations such as object detection and robot pose estimation.

\item \texttt{CreatePlanner}: The client provides hyperparameters of the planner, and the server creates a planner instance accordingly (\eg, POUCT planner in \texttt{pomdp\_py} \cite{pomdp-py-2020}).

\item \texttt{PlanAction}: The client requests the server to plan an action for an agent. An action is planned only if the last planned action has been executed successfully.

\item \texttt{ListenServer}: This is a bidirection streaming RPC that establishes a channel of communication of messages or status between the client and the server.
\end{itemize}

\subsection{Frames for Spot Finding Cat Underneath Couch}
\begin{figure}[H]
  \centering
  \includegraphics[width=\linewidth]{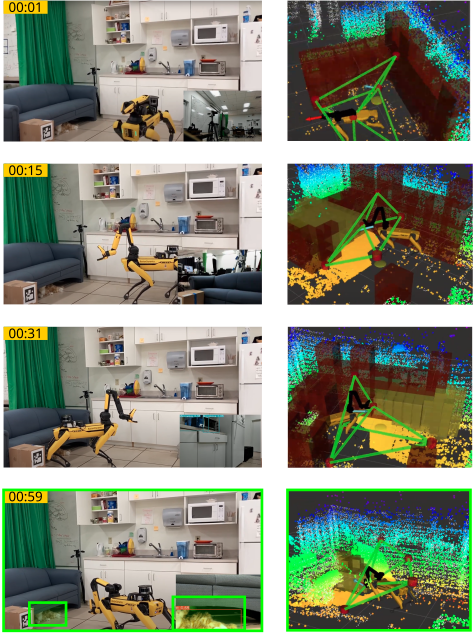}
  \caption{Sequence of frames from the search trial GenMOS enables Spot
    to find the toy cat hidden under the couch under one minute. Each internals visualization (right column) contains the
    octree belief (red boxes), point cloud, the view position graph, and the
    robot.}

  \label{fig:spot_cat_seq}
\end{figure}

\subsection{Example Configuration Parameters}

Below lists some parameters that the GenMOS server considers for configuration.

\begin{itemize}[leftmargin=*,noitemsep]
  \item\texttt{octree\_size}: Dimension of the octree covering the search region (\eg, 32 means the octree occupies a $32^3$ grid);
  \item\texttt{res} :  Resolution of a grid, \ie, length of the grid's side in meters (\eg, 0.1);
  \item\texttt{region\_size} :  Defines the dimensions of a box $(w,\ell,h)$ in meters (\eg, (4.0, 3.0, 1.5));
  \item\texttt{center} :  Defines the XYZ location (metric)of the search region's center (\eg, (-0.5, -1.65, 0.25)));
  \item\texttt{prior\_from\_occupancy}: Whether to use occupancy-based prior
  \item\texttt{occupancy\_fill\_height}: Whether to consider the space below obstacles as part of search space;
  \item\texttt{num\_nodes}: maximum number of view positions on graph (\eg, 10);
  \item\texttt{sep}: minimum separation between nodes (in meters) (\eg, 0.4);
  \item\texttt{inflation}: radius to blow up obstacles for view position sampling;
  \item\texttt{num\_sims}: number of samples for MCTS-based online POMDP planning.
  \end{itemize}
  \clearpage


\end{document}